\documentclass[conference]{IEEEtran}
\IEEEoverridecommandlockouts
\usepackage{cite}
\usepackage{amsmath,amssymb,amsfonts}
\usepackage{algorithmic}
\usepackage[linesnumbered,ruled,vlined,spanish,onelanguage]{algorithm2e}
\usepackage{graphicx}
\usepackage{textcomp}
\usepackage{xcolor}
\usepackage{multirow}
\usepackage{balance}
\usepackage{todonotes}
\usepackage{xspace}

\usepackage{soul}
\usepackage{todonotes}

\def\BibTeX{{\rm B\kern-.05em{\sc i\kern-.025em b}\kern-.08em
    T\kern-.1667em\lower.7ex\hbox{E}\kern-.125emX}}
\begin{document}
\title{Evaluation of Self-taught Learning-based Representations for Facial Emotion Recognition\\

}

\newif\ifauthors
\authorstrue

\ifauthors
\author{
    \IEEEauthorblockN{
        Bruna Delazeri\textsuperscript{1}, 
        Leonardo L. Veras\textsuperscript{1}, 
        Alceu de S. Britto Jr.\textsuperscript{1,2}, 
        Jean Paul Barddal\textsuperscript{1},
        Alessandro L. Koerich\textsuperscript{3}
    }
    \IEEEauthorblockA{
        \\ 
        \textsuperscript{1}Graduate Program in Informatics (PPGIa), Pontifícia Universidade Católica do Paraná (PUCPR), Curitiba (PR), Brazil
        \\
        Email: \{bruna.delazeri, alceu, jean.barddal\}@ppgia.pucpr.br
        \\
         \textsuperscript{2}State University of Ponta Grossa (UEPG), Ponta Grossa (PR), Brazil 
        \\
        \textsuperscript{3}École de Technologie Supérieure (ÉTS), Université du Québec, Montréal (QC), Canada
        \\
        Email: alessandro.koerich@etsmtl.ca
    }
}
\else 
\author{
    \IEEEauthorblockN{Anonymous Authors}
    \vspace*{3.5cm}
}
\fi

\maketitle

\begin{abstract}

This work describes different strategies to generate unsupervised representations obtained through the concept of self-taught learning for facial emotion recognition (FER). The idea is to create complementary representations promoting diversity by varying the autoencoders' initialization,  architecture, and training data. SVM, Bagging, Random Forest, and a dynamic ensemble selection method are evaluated as final classification methods. Experimental results on Jaffe and Cohn-Kanade datasets using a leave-one-subject-out protocol show that FER methods based on the proposed diverse representations compare favorably against state-of-the-art approaches that also explore unsupervised feature learning.  
\end{abstract}

\begin{IEEEkeywords}
Self-taught Learging, Autoencoder, Unsupervised Representation Learning, Dynamic Ensemble Selection.
\end{IEEEkeywords}

\section{Introduction}
To build a model that generalizes properly, supervised learning methods require large amounts of labeled training data that, in many real-world applications, are scarce, difficult, and expensive to obtain. 
Consequently, this usually became a restriction in both computer vision and machine learning tasks. As a result, knowledge transfer between tasks emerged as a new learning framework to deal with the lack of labeled data. Self-taught learning (STL) is a particular transfer learning method that exploits data with a different distribution than the target problem \cite{raina}. In opposition to transfer learning, STL does not require labeled data from an auxiliary domain, as it learns representations without labeling and considering different data distributions. 

The use of unlabeled data has been the main advantage highlighted by authors in STL studies \cite{raina}. The rationale behind STL has foundations borrowed from natural human learning. It is believed to rely on unlabeled data, which helps provide a solid foundation for high-level learning, thus generating more significant discriminative power \cite{raina}. In computational terms, STL is relevant when:
(i) there is little labeled data for training \cite{BETTGE}; or
(ii) despite having sufficiently enough labeled data, using examples from outside the classes of interest improves the learning process due to a greater generalization power \cite{BASTIEN}.

This paper is mainly concerned with STL and its application in facial emotion recognition (FER). Facial expression is a natural and very significant way for human beings to convey their emotional state in the communication process. 
Researchers have investigated FER systems due to the various applications that benefit from their use, such as human interaction \cite{ZHANG}, disease diagnosis \cite{ZHOU}, virtual reality \cite{HICKSON}, augmented reality \cite{CHEN}, and driver fatigue monitoring \cite{QIAO}. FER is a task in which STL is feasible since there are several publicly available datasets, but the amount of labeled data for training robust models is scarce. Besides, the data variability is challenging due to the presence of people of different ages, skin colors, cultures, and genders \cite{sinno}. Consequently, we believe STL is an appropriate approach for developing a robust FER system since it bypasses the necessity of having large datasets with diverse and labeled data.


In such a context, this paper proposes strategies to generate unsupervised representations using the concept of STL with a focus on diversity. For such an aim, convolutional autoencoders (CAE) are trained considering strategies that promote diversity, such as different model initializations, architectures, and training data. The rationale is to investigate how complementary learned representations can contribute to the performance of a FER solution created using a small dataset. 

We have two main research questions (RQs). The first one (RQ1) is related to the idea of exploring diverse unsupervised representations: \textit{"Could the use of a pool of unsupervised representations contribute to the performance of a supervised classification model?} The second question (RQ2) concerns the strategy used to promote diversity: \textit{Which could be the strategy used to promote diversity when creating a pool of unsupervised representations?}

To answer these RQs, we investigate four strategies to generate a diverse pool of representations. First, the complementary of their members are evaluated, considering the performance of ensembles (Bagging and Random Forest) and a dynamic ensemble selection method (KnoraU) \cite{Ko} produced based on them. Experiments on two FER databases using the leave-one-subject-out (LOSO) protocol have shown that promising results can be achieved by combining distinct representations. Moreover, the best strategies are based on varying the architecture of CAE.

This paper is organized as follows.
Section \ref{relatedwork} brings forwards works related to our research.
Section \ref{proposedmethod} describes the proposed method for automatic representation generation using STL.
Section \ref{results} presents our experimental results and a comparison with the state-of-the-art. Finally, Section \ref{conclusion} concludes this work and states envisioned future work.

\section{Related Works} \label{relatedwork}
Proposed by \cite{raina}, STL, also described as \textit{unsupervised transfer} or \textit{transfer of learning of unlabeled data}, is a machine learning framework that requires little human supervision. As a result, several authors have used STL in different classification applications such as audio \cite{markov}, text \cite{han,QURESHI}, image  \cite{Sheng}, and sensor data \cite{Peilin}.

Initially, such a learning technique used out-of-distribution examples as a source of unlabeled data, showing positive effects in scenarios where labeled data was limited. The results presented in the seminal work of \cite{raina}, who used a sparse surface coding for representation learning, showed that the relative gain of STL decreases as the number of labeled examples increases. However, studies using deep architectures have shown that such a positive effect is achieved even in a scenario with a large number of labeled examples \cite{ERHAN}. Another relevant characteristic of STL is that the deep layers of neural networks have hierarchically distributed characteristics that can be shared between tasks and data distributions.

Concerning the application of STL for FER, two very interesting works can be highlighted. The first \cite{LONG} describes a video FER system. An unsupervised feature learning method based on independent component analysis (ICA) learns spatiotemporal filters from natural videos, which are used to represent the images of facial expressions. They reported the area under the ROC curve for each of six classes of emotions present in the Cohn-Kanade database, ranging from  0.69 to 0.89.

The second one is the work presented in \cite{BHANDARI} that compares representations based on STL and transfer learning for FER. They trained base models on the CIFAR10 image dataset and applied them to the JAFFE expression recognition dataset. Their STL approach using a sparse autoencoder for feature extraction and a final CNN that receives the weights learned in the unsupervised step achieved an accuracy of 56.45\%.

Without considering STL but exploring diversity to create ensembles of CNNs, the authors in \cite{RENDA} investigate different strategies for inducing diversity in an ensemble of CNNs applied to FER. The results on the FER2013 dataset showed that seed variation yielded the best recognition results while variations on the pre-training process of their CNNs achieved the best run-time performance.

Inspired by the works of \cite{BHANDARI} and \cite{RENDA}, our idea is to investigate whether a diverse pool of representations learned using STL can bring some improvement in the accuracy of FER created using small FER datasets. We expect that unsupervised representations can generate ensembles composed of diverse members.

\section{Proposed Method} \label{proposedmethod}

\begin{figure}
    \centering
    \includegraphics[scale=0.6]{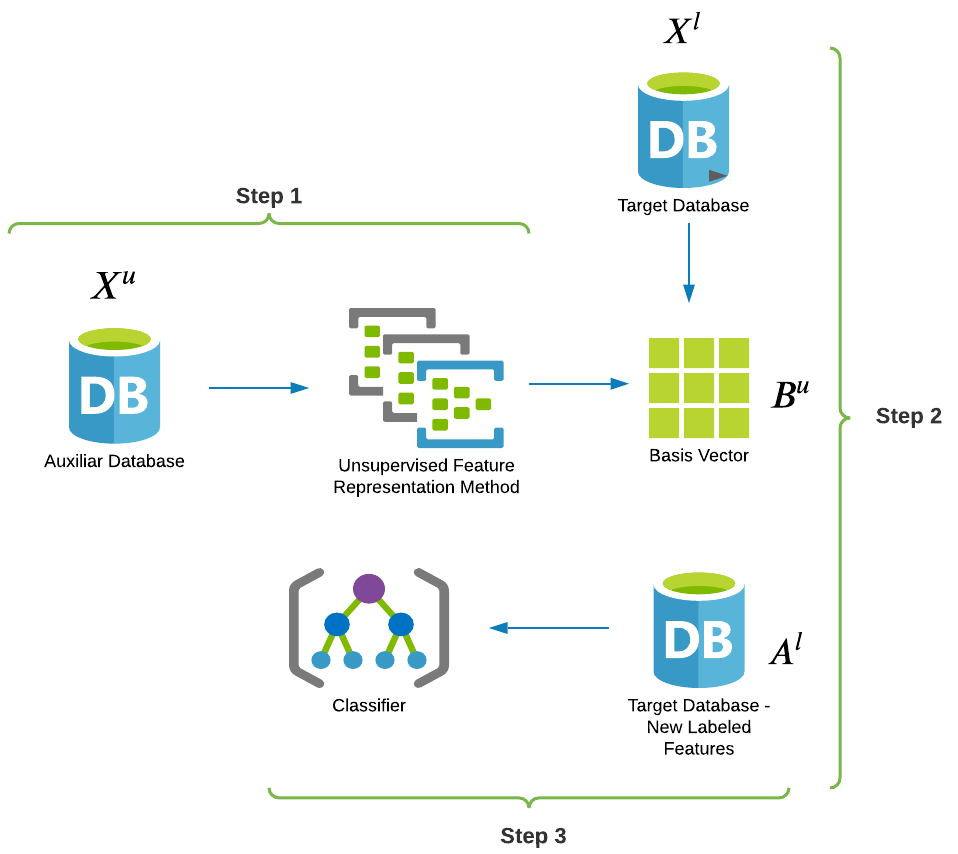}
    \caption{Proposed method based on the three STL steps: (1) unsupervised representation learning; (2) extraction of features from labeled data of the target domain; (3) supervised learning to create a classifier (or regressor) for the target domain represented by the feature vectors.}
    \label{FigSTL}
\end{figure}

This section describes the proposed method for generating pools of unsupervised representations using the concept of STL, detailing the different strategies evaluated to promote diversity. The rationale behind it is to investigate the impact of combining different but complementary unsupervised learned representations for the problem of FER. Fig.~\ref{FigSTL} presents a general overview of the proposed method, which is organized according to the three main steps of STL \cite{HUANG2009}:

\begin{itemize}
    \item \textbf{Representation Learning (Step 1)}: High-level representation is learned through unlabeled data, which does not necessarily present the same distribution as the labeled data of the target domain.
    \item \textbf{Feature Building (Step 2)}: Feature vectors are extracted from the labeled data of the target domain using the representation learned in Step 1.
    \item \textbf{Training a Classifier (Step 3)}: The feature vectors extracted in Step 2 are used to train a classifier (monolithic or ensemble).
\end{itemize}

\subsection{Representation Learning (Step 1)}
This step learns in an unsupervised way high-level representations. To learn them from unlabeled data, given an input \({ x }_{ i }^{ u }\in { R }^{ d }\) the unsupervised representation learning method tries to find patterns (base vectors) \({ B }^{ u }=\{ { b }_{ 1 }^{ u },{ b }_{ 2 }^{ u },...{ b }_{ k }^{ u }\} \in { { { R }^{ d\times K } } }\) such that each individual entry \({ x }_{ i }^{ u }\) can be represented using a combination of a few basic functions as denoted in Eq.~\eqref{eq:rep}.

	\begin{equation}
    { x }_{ i }^{ u }=\sum _{ k=1 }^{ k }{ { a }_{ i,k }^{ u } } { b }_{ k }^{ u }
    \label{eq:rep}
    \end{equation}

\noindent where \({ a }_{ i,k }^{ u }\) is the combination of the coefficients of \({ x }_{ i }^{ u }\), called activations and \({ b }_{ k }^{ u }\) are the base functions, i.e. high-level features. To allow that only some base vectors are used for an input \({ x }_{ i }^{ u }\), only some activation values will be nonzero. Taking into account a set of unlabeled data, Eq.~\eqref{eq:rep} can be rewritten as:

	\begin{equation}
	    { X }^{ u }={ B }^{ u }{ A }^{ u }
	    \label{eq:repm}
	\end{equation}
	
\noindent where \({ X }^{ u }=\{ { x }_{ 1 }^{ u },{ x }_{ 2 }^{ u },...,{ x }_{ n } ^{ u }\} \in { R }^{ d\times N }\) is the product of two matrices \({ B }^{ u }=\{ { b }_{ 1 }^{ u }, { b }_{ 2 }^{ u },...{ b }_{ k }^{ u }\} \in { { { R }^{ d\times K} } }\) and \({ A }^{ u }=\{ { a }_{ 1 }^{ u },{ a }_{ 2 }^{ u },...{ a }_{ N }^{ u }\} \in { { { R }^{ K \times N } } }\), each \({ a }_{ i }^{ u }\) represents the coefficient vector of the given vector \({ x }_{ i }^ {u}\). This equation decomposes the data matrix \({ X }^{ u }\) into two matrices \({ A }^{ u }\), known as the activation matrix and \({ B }^{ u }\), known as a dictionary array.

\begin{figure*}
    \centering
    \includegraphics[scale=0.6]{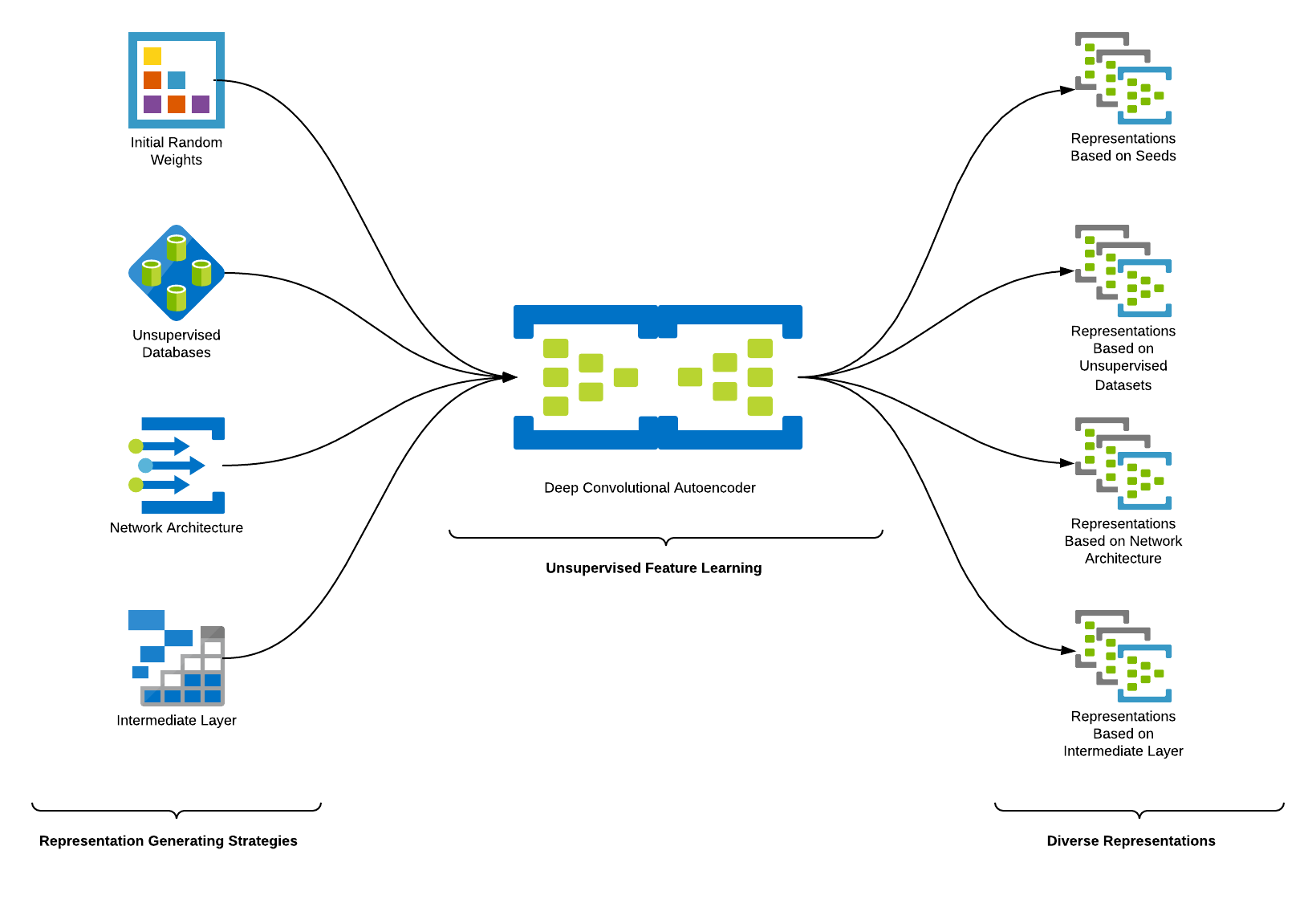}
    \caption{Automatic generator of representations based on different strategies.}
    \label{Fig_gerador}
\end{figure*}

We can find in the literature robust evidence that greater diversity is highly correlated with the increase in supervised CNN-based ensembles accuracy \cite{RENDA, Diver}. Thus, we decide to explore diversity to generate our ensemble of unsupervised representations as illustrated in Fig.~\ref{Fig_gerador}. The proposed algorithm uses a convolutional autoencoder (CAE) and different strategies to vary specific parameters, as follows:

\begin{itemize}
 \item \textbf{Random Seed}: here, different representations are generated by varying the distribution of weights during the CAE initialization process. The architecture is the same from one CAE to another, but the seed of the initialization process used differs. The input is the number of representations ($R$) to be generated with different seeds. The algorithm randomly selects a seed from the range $[0,1000]$ at each iteration. 
   \item \textbf{Training Dataset}: in this case, the same CAE is trained on different unlabeled datasets. The promotion of diversity is clear but is also an opportunity to evaluate the use of datasets far from the target domain. In this direction, we have considered four datasets of different fields during our experiments. 
    \item \textbf{Network Architecture}: here, we explore the generation of representations using different CAE architectures. We must define the network depth (number of layers, $D$), the filters used in each one, and the size ($I$) of the intermediate (latent) layer. The generator will create the first architecture with a depth equal to the number of defined layers  ($D$), then the second will have ($D-1$) up to the last architecture with depth $D=1$, which has one input, one intermediate, and one output layer (basic structure of a CAE). The representations generated here differ on the CAE architectures' depth but use the same number of neurons ($I$) in the latent layer. 
\item \textbf{Latent Vector}: in this strategy, we use the same CAE architecture, i.e., the encoder and decoder are the same for all generated representations. The diversity is obtained by varying the number of neurons in the middle layer of the network, named latent vector. The input is a list of $L$ different sizes for the intermediate layer. Thus, $L$ representations will be created with the sizes defined in that list.
\end{itemize}


\begin{figure}
    \centering
    \includegraphics[scale=0.6]{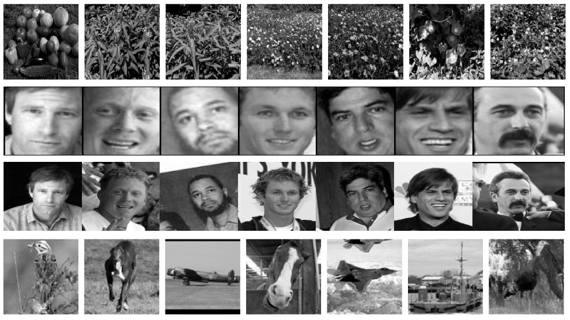}
    \caption{Image samples. In the first line images of the Kyoto database. In the second line, images of the LFW database with the removal of the image background. The third row has integral images of the LFW database. The fourth line illustrates the STL-10 database samples.}
    \label{fig_bases}
\end{figure}

\subsection{Feature Building (Step 2)}
According to the STL approach, after having learned the dictionary \({ B }^{ u }\), from the input data \({ X }^{ u }\), it is used to obtain the activations of the data labeled \({ X }^{ l }\). Thus, the vectors labeled \({ x }_{ i }^{ l }\) can be represented as a combination of some base functions:

	\begin{equation}
	    { X }^{ l }={ B }^{ u }{ A }^{ l }
	\end{equation}
	
\noindent where \({ A }^{ l }=\{ { a }_{ 1 }^{ l },{ a }_{ 2 }^{ l },...{ a }_{ M }^ { lu }\} \in { { { R }^{ KxM } } }\) is the activation matrix corresponding to the labeled data. This can be considered a new way of representing \({ X }^{ l }\), where it is possible to assign to the original class \({ y }_{ i }\) each activation vector \({ a } _{ i }^{ l }\) , and then obtain a new representation for the target labeled data, which can be used to build some classifier in a supervised way.

Each representation learned in the previous step (Diverse Representation on \ref{FigSTL}) is used to extract feature vectors from the target dataset. Thus, the unsupervised learned representations generate new feature sets using the different techniques described.

\begin{figure}
    \centering
    \includegraphics[scale=0.45]{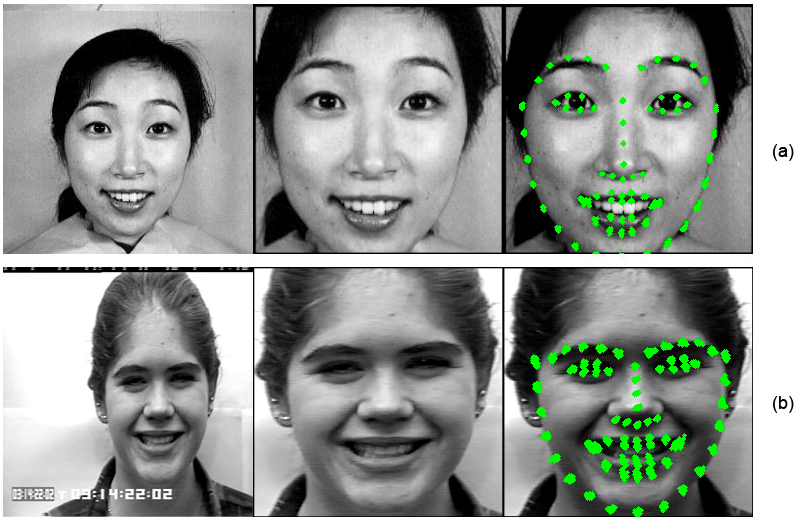}
    \caption{Example of an original image, face detection, and landmark extraction of the Jaffe dataset (a) and CK dataset (b).}
    \label{figimgscrop}
\end{figure}

In this step, before extracting the features from the FER images, we pre-process them, cutting only the region of the area of interest (face) and selecting the reference points. Figure \ref{figimgscrop} illustrates the pre-processing applied to a sample image of Jaffe and CK datasets. To detect and cut out the face area of the image, we used the Viola-Jones face detection method. The reference points located in the cropped image are used to align the images, leaving them in the same position. 

The pre-processed image is the input of each learned CAE, which use their encoder weights \({ w }_{ 1 }\) and bias \(b\), to extract features \(a={ h }_{ w1,b }({ x }^{ l })\) from the labeled data \(({ X }^{ l },Y) \). Every learned representation is finally transformed using Principal Component Analysis (PCA) \cite{Jolliffe} to reduce its dimensionality, while decorrelating the learned features. We use this new representation \(a\), with the label vector \(Y\), to perform the third step of the STL approach, the classification.

\subsection{Training a Classifier (Step 3)}

At this point, we have the target datasets represented by $R$ different strategies to generate supervised models. Thus, we evaluated a monolithic approach (SVM), two ensemble learning algorithms (Bagging and Random Forest - RF), and a dynamic ensemble selection method (Knora Union - KnoraU). Table \ref{table_parameters} presents the parameters used for each technique. The motivation for using a single classifier, ensembles, and a dynamic selection method is to investigate the impact of the generated diversity.


\section{Experimental Results} \label{results}
The target datasets are the Japanese Female Facial Expression (JAFFE) \cite{Lyons1998} and the Cohn-Kanade (CK). JAFFE is a laboratory-controlled image database that has 213 images of 10 subjects (Japanese female models), with six basic facial expressions (happiness, anger, disgust, fear, sadness, and surprise) plus a neutral one. On JAFFE, all ten subjects have one or more images for each class. Another motivation for using JAFFE is that its images are grayscale, making them remarkably different from those used to learn our representations. CK is a laboratory-controlled database widely used to evaluate FER systems. The base contains 523 sequences from 123 subjects, 327 images of 118 subjects are labeled with seven basic facial expression (anger, contempt, disgust, fear, joy, sad and surprise). On this database not all subjects have images in all classes.

The following datasets were used to generate the unsupervised representations (auxiliary datasets):

\begin{itemize}
    \item Kyoto Natural Images: The dataset contains 62 natural images of 256$\times$200 pixels, used by several other works that apply the concepts of STL, thanks to its large variability. Our experimental protocol's images belong to a domain (FER) far from that dataset. 
    \item LFW: is a publicly available database of face photographs used for face verification, also known as peer matching. The dataset has 13,233 images of faces collected on the web from 5,749 people. Unlike Kyoto, LFW was selected for our protocol because its images belong to a domain related to the FER problem.
    \item LFW-Face: Similar to the LFW dataset, LFW-Face is close to our target domain. However, it uses only the bounding box of the face instead of the full image with the background.
    \item STL-10: It is an image recognition database frequently used to develop unsupervised feature learning and deep models. The database has color images with 96x96 pixels categorized into ten classes. Here the labels were ignored. 
\end{itemize}

In our experimental protocol, we use the leave-one-subject-out (LOSO) cross validation. This protocol divides the database so that a subject in the test dataset cannot be in training. Suppose we have a dataset with $N$ subjects, therefore, for each \textit{fold}, one subject will be used as a test and the others for training. In such a protocol, when using the KNORAU dynamic ensemble selection, a subset of data (validation set) is separated from the training set. Such a validation set is necessary since KNORAU needs to compute the competence of each classifier.It is also essential to notice that all available images were used when training the CAEs with a specific auxiliary dataset. Moreover, in all experiments, except in Experiment 2 (varying the datasets), Kyoto was used as the auxiliary dataset. 


A set of five experiments were performed to answer our research questions. One to evaluate each of four strategies used to generate the unsupervised representations and a last one to combine all the results. Table \ref{table_parameters} reports the parameter settings of the algorithms used to generate the supervised models in our experiments. The CAE parameters used in each experiment are shown in Table \ref{parameters_autoencoder}. The inducers were executed using the implementations from the scikit-learn Python library\footnote{Available at http://scikit-learn.org/}, The CAEs were implemented using Keras\footnote{Available at https://keras.io/} with TensorFlow, while the dynamic ensemble selection method KnoraU was implemented using Deslib \footnote{Available at https://deslib.readthedocs.io/en/latest/api.html}.

\begin{table}[htbp]
\centering
\caption{Parameter settings of algorithms used in Step 3. Classifiers: single (SVM), Ensembles (BG: Bagging with Decision Tree as base classifier, RF: Random Forest), KnoraU (DT: pool of trees; RF: Random Forest). }
\begin{tabular}{l|ll} \hline
\multicolumn{1}{c|}{\textbf{Algorithm}}                              & \multicolumn{2}{c}{\textbf{Parameters}}                                                                 \\ \hline
\multirow{4}{*}{SVM}                                                  & \multicolumn{1}{l|}{Kernel}                    & Linear                                                  \\
                                                                      & \multicolumn{1}{l|}{Penalty Parameter (C)}     & 1e-6                                                    \\
                                                                      & \multicolumn{1}{l|}{Class Weight}              & Balanced                                                \\
                                                                      & \multicolumn{1}{l|}{Probability}               & True                                                    \\ \hline
\multirow{4}{*}{\begin{tabular}[c]{@{}l@{}}BG with DT\end{tabular}} & \multicolumn{1}{l|}{Max Depth}                 & 10                                                      \\
                                                                      & \multicolumn{1}{l|}{Tree Max Features}              & sqrt                                                    \\
                                                                      & \multicolumn{1}{l|}{Number of Base Estimators} & 100                                                     \\
                                                                      & \multicolumn{1}{l|}{\% of Training Samples }         & 1.0                                                     \\ \hline
\multirow{3}{*}{RF}                                                   & \multicolumn{1}{l|}{Max Depth}                 & 10                                                      \\
                                                                      & \multicolumn{1}{l|}{Number of Trees}           & 100                                                     \\
                                                                      & \multicolumn{1}{l|}{Oob\_score}                & True                                                    \\ \hline
\multirow{2}{*}{KNORA}                                                & \multicolumn{1}{l|}{Number of neighbors}       & 7                                                       \\
                                                                      & \multicolumn{1}{l|}{pool\_classifiers}         & \begin{tabular}[c]{@{}l@{}}Bagging DT\\ RF\end{tabular} \\ \hline
\multirow{2}{*}{Stacking}                                                & \multicolumn{1}{l|}{Meta classifier}       & Logistic Regression                                                       \\
                                                                      & \multicolumn{1}{l|}{Solver}         & \textit{lbfgs} \\ 
                                                                      
                                                                      & \multicolumn{1}{l|}{Penalty parameter $C$}         & 1.0 \\
                                                                      
                                                                      \hline

\end{tabular}
\label{table_parameters} 
\end{table}

\begin{table}[htbp]
\centering
\caption{Parameter Settings used in Step 2 - Feature Building, CAE architectures and PCA components}
\begin{tabular}{l|l|l} \hline
\multicolumn{1}{c}{\textbf{Algorithm}}                                                        & \multicolumn{2}{c}{\textbf{Parameters}}                                                                                          \\ \hline
\multirow{9}{*}{\begin{tabular}[c]{@{}l@{}}CAE - \\ Seeds\\ Unlabeled \\ Datasets\end{tabular}} & Input Length       & 96X96X1                                                                                                     \\
                                                                                              & Activation         & ReLU/Softmax                                                                                                \\
                                                                                              & Lattent Vector     & $I$ = 2500                                                                                                  \\
                                                                                              & Epochs             & 20                                                                                                          \\
                                                                                              & Loss               & mse                                                                                                         \\
                                                                                              & Optimizer          & SGD                                                                                                         \\
                                                                                              & Filter Convolution & 3x3                                                                                                         \\
                                                                                              & Filters            & 16, 32, 64, 128                                                                                             \\
                                                                                              & Network Depth  $D$ & \begin{tabular}[c]{@{}l@{}}5 encoder\\ 1 Lattent Vector\\ 5 decoder\end{tabular}                            \\ \hline
\begin{tabular}[c]{@{}l@{}}CAE - \\ Network \\ Architecture\end{tabular}                         & Network Depth      & \begin{tabular}[c]{@{}l@{}}$D$ = 5\\ $N$ = \{$D$, $D-1$...1\}\end{tabular}                                  \\ \hline
\begin{tabular}[c]{@{}l@{}}CAE - \\ Lattent Vector\end{tabular}                               & Lattent Vector     & \begin{tabular}[c]{@{}l@{}}$I$ = {[}150, 200, 250, 300, 400, \\ 500, 1000, 1500, 2000, 2500{]}\end{tabular} \\ \hline
\begin{tabular}[c]{@{}l@{}}PCA  \\ \end{tabular}                               & \# of components     & \begin{tabular}[c]{@{}l@{}}{150}\end{tabular} \\ \hline
\end{tabular}  
\label{parameters_autoencoder}
\end{table}

\subsection{Experiment 1 - Varying the Random Seeds}
Tables \ref{table_seeds} and \ref{table_seeds_CK} show the results obtained with our first strategy considering $R=10$ representations using different seeds to initialize the CAEs that were applied to represent each target dataset, JAFFE and CK, respectively. The CAEs were trained here using the Kyoto dataset. We can see that the fusion of all representations in the last three lines of these tables show a significant improvement no matter the classification approach used (SVM, Ensembles, or KnoraU). The best result for JAFFE was 56.60\%, observed when combining the output of the RFs trained on each representation using the product rule, while the best outcome for CK was 84.39\% observed when combining the output of the SVMs trained on each representation using the stacking fusion strategy.


\begin{table}[htbp]
\centering
\caption{Results of varying random seeds (10 representations) on JAFFE Database. Classifiers: single (SVM), Ensembles (BG: Bagging, RF: Random Forest), KnoraU (DT: pool of trees; RF: Random Forest). Fusion of all (sum, product and stacking) }
\begin{tabular}{l|l|ll|ll} \hline
\multicolumn{1}{c|}{} & \multicolumn{1}{c|}{} & \multicolumn{2}{c|}{Ensembles} &  \multicolumn{2}{c}{KnoraU}   \\                 
Repr.                & \multicolumn{1}{c|}{SVM}                     & \multicolumn{1}{c}{BG}                    & \multicolumn{1}{c|}{RF}                    & \multicolumn{1}{c}{DT} & \multicolumn{1}{c}{RF} \\ \hline
Seed 1               & 50.66                                    & 37.75                                   & 36.28                                   & 38.16                  & 35.42                  \\
Seed 2               & 29.83                                    & 38.84                                   & 37.82                                   & 40.11                  & 35.58                  \\
Seed 3               & 44.30                                    & 33.06                                   & 33.87                                   & 37.63                  & 33.47                  \\
Seed 4               & 34.27                                    & 33.22                                   & 31.82                                   & 34.53                  & 34.86                  \\
Seed 5               & 49.75                                    & 39.40                                   & 36.50                                   & 38.38                  & 34.60                  \\
Seed 6               & 52.90                                    & 40.67                                   & 38.54                                   & 39.67                  & 38.91                  \\
Seed 7               & 42.34                                    & 36.58                                   & 32.00                                   & 35.90                  & 34.77                  \\
Seed 8               & 48.34                                    & 43.64                                   & 39.61                                   & 40.25                  & 40.50                  \\
Seed 9               & 28.03                                    & 32.72                                   & 33.80                                   & 34.15                  & 34.79                  \\
Seed 10              & 35.33                                    & 34.01                                   & 37.78                                   & 37.30                  & 35.85                  \\ \hline
Sum                  & \textbf{52.15}                                    & 52.88                                   & 56.53                                   & 56.20         & 50.12                  \\
Product              & 51.74                                    & \textbf{54.35}                                   & \textbf{56.60}                                   & 56.20         & 50.12                  \\
Stacking             & 48.93                                    & 52.54                                   & 54.67                                   & \textbf{55.68}         & \textbf{54.79}       \\ \hline          
\end{tabular}
 \label{table_seeds} 
\end{table}

\begin{table}[htbp]
\centering
\caption{Results of varying random seeds (10 representations) on CK Database. Classifiers: single (SVM), Ensembles (BG: Bagging, RF: Random Forest), KnoraU (DT: pool of trees; RF: Random Forest). Fusion of all (sum, product and stacking) }
\begin{tabular}{l|l|ll|ll} \hline
\multicolumn{1}{c|}{} & \multicolumn{1}{c|}{} & \multicolumn{2}{c|}{Ensembles} &  \multicolumn{2}{c}{KnoraU}   \\                 
Repr.                & \multicolumn{1}{c|}{SVM}                     & \multicolumn{1}{c}{BG}                    & \multicolumn{1}{c|}{RF}                    & \multicolumn{1}{c}{DT} & \multicolumn{1}{c}{RF} \\ \hline
Seed 1               & 84.22                                    & 61.01                  & 60.15                  & 65.56                  & 60.90                  \\
Seed 2               & 82.10                                    & 57.90                  & 56.35                  & 58.23                  & 58.96                  \\
Seed 3               & 80.40                                    & 63.98                  & 59.95                  & 65.42                  & 60.12                  \\
Seed 4               & 84.06                                    & 66.51                  & 63.95                  & 66.93                  & 65.53                  \\
Seed 5               & 81.56                                    & 63.41                  & 60.15                  & 63.20                  & 61.07                  \\
Seed 6               & 78.09                                    & 63.03                  & 62.34                  & 64.97                  & 62.16                  \\
Seed 7               & 80.08                                    & 66,86                  & 64.46                  & 66.41                  & 67.00                  \\
Seed 8               & 82.71                                    & 64.06                  & 59.80                  & 64.84                  & 59.92                  \\
Seed 9               & 81.10                                    & 64.60                  & 62.8                   & 65.12                  & 62.86                  \\
Seed 10              & 80.93                                    & 60.50                  & 56.56                  & 59.61                  & 58.26                  \\ \hline
Sum                  & 83.38                                    & 71.31                  & 70.83                  & 70.83                  & 66.04                  \\
Product              & 83.03                                    & 71.63                  & 69.77                  & 69.77                  & 66.68                  \\
Stacking             & \textbf{84.39}                           & \textbf{77.48}         & \textbf{72.64}         & \textbf{72.64}         & \textbf{70.98}                    \\
 \hline          
\end{tabular}
 \label{table_seeds_CK} 
\end{table}

\subsection{Experiment 2 - Varying the Training Datasets}
Tables \ref{table_databases} and \ref{table_databases_CK} present the results obtained by the representations generated with the four different auxiliary datasets previously described, considering JAFFE and CK datasets, respectively. Concerning the proximity of the auxiliary dataset to the target domain, we observed that the datasets far from the target domain presented the best results in 6 over ten experiments. Similar to the first set of experiments, we can see that the fusion of all representations in the last three lines of these tables show a significant improvement no matter the classification approach used. The best result for JAFFE was 61.69\%, observed when combining the output of the SVMs trained on each representation using the stacking strategy, while the best outcome for CK was 86.92\% observed using the same classification approach.

\begin{table}[htbp]
\caption{Results of varying unlabeled datasets (4 representations) on JAFFE Database. Classifiers: single (SVM), Ensembles (BG: Bagging, RF: Random Forest), KnoraU (DT: pool of trees; RF: Random Forest). Fusion of all (sum, product and stacking) }
\centering
\begin{tabular}{l|l|ll|ll} \hline
\multicolumn{1}{c|}{} & \multicolumn{1}{c|}{} & \multicolumn{2}{c|}{Ensembles} &  \multicolumn{2}{c}{KnoraU}   \\                 
Repr.                & \multicolumn{1}{c|}{SVM}                     & \multicolumn{1}{c}{BG}                    & \multicolumn{1}{c|}{RF}                    & \multicolumn{1}{c}{DT} & \multicolumn{1}{c}{RF} \\ \hline
Kyoto    & 60.35                                   & 41.08                                   & 35.55                                   & 43.40                  & 37.57                  \\
LFW-Face & 59.44                                    & 44.22                                   & 38.39                                   & 41.48                  & 35.70                  \\
LFW      & 58.94                                    & 40.06                                   & 38.71                                   & 37.71                  & 40.12                  \\
STL-10   & 55.21                                    & 34.94                                   & 35.04                                   & 35.51                  & 34.80                  \\ \hline
Sum      & 59.40                           & 50.27                                   & 45.47                                   & 51.31                  & 46.92                  \\
Product  & 59.85                           & 50.33                                   & 46.87                                   & 51.85                  & \textbf{47.42}                  \\
Stacking & \textbf{61.69}                           & \textbf{51.72}                                   & \textbf{47.87}                                   & \textbf{53.60}                  & 47.08  \\ \hline                
\end{tabular}
 \label{table_databases} 
\end{table}

\begin{table}[htbp]
\caption{Results of varying unlabeled datasets (4 representations) on CK Database. Classifiers: single (SVM), Ensembles (BG: Bagging, RF: Random Forest), KnoraU (DT: pool of trees; RF: Random Forest). Fusion of all (sum, product and stacking) }
\centering
\begin{tabular}{l|l|ll|ll} \hline
\multicolumn{1}{c|}{} & \multicolumn{1}{c|}{} & \multicolumn{2}{c|}{Ensembles} &  \multicolumn{2}{c}{KnoraU}   \\                 
Repr.                & \multicolumn{1}{c|}{SVM}                     & \multicolumn{1}{c}{BG}                    & \multicolumn{1}{c|}{RF}                    & \multicolumn{1}{c}{DT} & \multicolumn{1}{c}{RF} \\ \hline
Kyoto    & 84.19                                    & 64.90                                   & 59.95                                   & 64.45                  & 61.85                  \\
LFW-Face & 85.08                                    & 63.74                                   & 58.99                                   & 63.16                  & 60.19                  \\
LFW      & 79.94                                    & 65.11                                   & 59.01                                   & 64.66                  & 59.77                  \\
STL-10   & 83.07                                    & 65.84                                   & 64.80                                   & 66.69                  & 65.72                  \\ \hline
Sum      & 86.22                                    & 69.87                                   & 64.26                                   & 70.01        & 65.22                  \\
Product  & 86.01                                    & 69.02                                   & 64.83                                   & \textbf{70.81}         & 65.01                  \\
Stacking & \textbf{86.92}                           & \textbf{70.31}                          & \textbf{68.89}                          & 69.73                  & \textbf{69.08}        \\ \hline               
\end{tabular}
 \label{table_databases_CK} 
\end{table}

\subsection{Experiment 3 - Varying the CAE Architecture}
In this experiment we have evaluated the use of different CAEs architectures. Tables \ref{table_architecture} and \ref{table_architecture_CK} show the results for JAFFE and CK datasets, respectively. We also can observe a significant contribution by combining the created representations. The best result for JAFFE was 59.67\%, observed when combining the output of the SVMs trained on each representation using the stacking strategy, while the best outcome for CK was 87.21\% observed using the same base classifier but with product rule. For the size of the architecture ($D$), we observed that the best results were achieved with architectures containing $D>=3$ layers (6 over 10 experiments).

\begin{table}[htbp]
\centering
\caption{Results of varying networks architectures (5 representations) on JAFFE Database. Classifiers: single (SVM), Ensembles (BG: Bagging, RF: Random Forest), KnoraU (DT: pool of trees; RF: Random Forest). Fusion of all (sum, product and stacking)}
\begin{tabular}{l|l|ll|ll} \hline
\multicolumn{1}{c|}{} & \multicolumn{1}{c|}{} & \multicolumn{2}{c|}{Ensembles} &  \multicolumn{2}{c}{KnoraU}   \\                 
Repr.                & \multicolumn{1}{c|}{SVM}                     & \multicolumn{1}{c}{BG}                    & \multicolumn{1}{c|}{RF}                    & \multicolumn{1}{c}{DT} & \multicolumn{1}{c}{RF} \\ \hline
Arch-1                    & 53.41                                    & 40.15                                   & 40.11                                   & 40.15                  & 37.95                  \\
Arch-2                    & 52.81                                    & 41.09                                   & 36.17                                   & 44.36                  & 35.45                  \\
Arch-3                    & 56.39                                    & 37.39                                   & 29.88                                   & 34.18                  & 36.46                  \\
Arch-4                    & 47.53                                    & 30.82                                   & 36.69                                   & 31.79                  & 38.08                  \\
Arch-5                    & 50.15                                    & 41.69                                   & 37.48                                   & 34.59                  & 40.25                  \\ \hline
Sum                       & 59.26                           & \textbf{51.82}                                   & 49.95                                   & 50.00                  & 50.36                  \\
Product                   & 58.76                           & 51.84                                   & 50.53                                   & 50.45                  & 49.91                  \\
Stacking                  & \textbf{59.67}                           & 50.41                                   & \textbf{50.96}                                   & \textbf{51.34}                  & \textbf{53.79}        \\ \hline         
\end{tabular}
 \label{table_architecture}
\end{table}

\begin{table}[htbp]
\centering
\caption{Results of varying networks architectures (5 representations) on CK Database. Classifiers: single (SVM), Ensembles (BG: Bagging, RF: Random Forest), KnoraU (DT: pool of trees; RF: Random Forest). Fusion of all (sum, product and stacking)}
\begin{tabular}{l|l|ll|ll} \hline
\multicolumn{1}{c|}{} & \multicolumn{1}{c|}{} & \multicolumn{2}{c|}{Ensembles} &  \multicolumn{2}{c}{KnoraU}   \\                 
Repr.                & \multicolumn{1}{c|}{SVM}                     & \multicolumn{1}{c}{BG}                    & \multicolumn{1}{c|}{RF}                    & \multicolumn{1}{c}{DT} & \multicolumn{1}{c}{RF} \\ \hline
Arch-1                              & 84.94                                    & 69.11                                   & 67.07                                   & 69.85                  & 67.24                  \\
Arch-2                              & 86.56                                    & 69.68                                   & 61.48                                   & 69.54                  & 65.22                  \\
Arch-3                              & 85.93                                    & 70.14                                   & 65.42                                   & 69.43                  & 67.11                  \\
Arch-4                              & 86.75                                    & 70.40                                    & 66.63                                   & 70.48                  & 65.83                  \\
Arch-5                              & 87.55                                    & 65.53                                   & 66.83                                   & 65.53                  & 68.12                  \\ \hline
Sum                                 & 87.20                                    & 74.12                                   & 70.76                                   & 73.48                  & 70.31                  \\
Product                             & \textbf{87.21}                           & 76.94                                   & 69.43                                   & \textbf{74.40}         & 70.31                  \\
Stacking                            & 86.85                                    & \textbf{78.20}                          & \textbf{73.54}                          & \textbf{74.40}         & \textbf{72.83}         \\ \hline
\end{tabular}
 \label{table_architecture_CK}
\end{table}

\subsection{Experiment 4 - Varying the Size of the Latent Vector}
The results of our last strategy are shown in Tables \ref{table_hidden} and \ref{table_hidden_CK}. The same network architecture is used with different intermediate layer sizes ($I$). The best result for JAFFE was 62.26\%, observed when combining the output of the SVMs trained on each representation using the stacking strategy, while the best outcome for CK was 86.99\% observed using the same base classifier but with product rule. For the size of the latent vector ($I$), we observed that the best results were achieved with $I >= 1500$ (7 over 10 experiments).

\begin{table}[htbp]
\centering
\caption{Results of varying latent vector size (10 representations) on JAFFE Database. Classifiers: single (SVM), Ensembles (BG: Bagging, RF: Random Forest), KnoraU (DT: pool of trees; RF: Random Forest). Fusion of all (sum, product and stacking)}
\begin{tabular}{l|l|ll|ll} \hline
\multicolumn{1}{c|}{} & \multicolumn{1}{c|}{} & \multicolumn{2}{c|}{Ensembles} &  \multicolumn{2}{c}{KnoraU}   \\                 
Repr.                & \multicolumn{1}{c|}{SVM}                     & \multicolumn{1}{c}{BG}                    & \multicolumn{1}{c|}{RF}                    & \multicolumn{1}{c}{DT} & \multicolumn{1}{c}{RF} \\ \hline
150      & 58.92                                    & 34.29                                   & 39.87                                   & 32.33                  & 37.91                  \\
200      & 59.85                                    & 41.68                                   & 36.59                                   & 39.24                  & 37.28                  \\
250      & 34.90                                    & 34.11                                   & 36.03                                   & 36.48                  & 34.41                  \\
300      & 59.92                                    & 33.53                                   & 33.25                                   & 38.82                  & 34.24                  \\
400      & 46.57                                    & 32.74                                   & 31.79                                   & 33.60                  & 37.08                  \\
500      & 59.96                                    & 42.77                                   & 41.64                                   & 44.37                  & 41.73                  \\
1000     & 41.46                                    & 37.28                                   & 32.21                                   & 32.86                  & 30.32                  \\
1500     & 50.49                                    & 43.54                                   & 38.77                                   & 44.84                  & 41.59                  \\
2000     & 46.09                                    & 37.81                                   & 33.27                                   & 41.14                  & 34.71                  \\
2500     & 51.33                                    & 41.66                                   & 44.51                                   & 42.14                  & 39.99                  \\ \hline
Sum      & 61.23                           & 59.52                                   & 56.25                                   & \textbf{59.83}                  & 57.64                  \\
Product  & 60.23                           & 59.46                                   & 57.66                                   & 59.33                  & 57.64                  \\
Stacking & \textbf{62.26}                           & \textbf{60.53}                                   & \textbf{58.09}                                   & 59.33                  & \textbf{58.86}            \\ \hline     
\end{tabular}
 \label{table_hidden}
\end{table}

\begin{table}[htbp]
\centering
\caption{Results of varying latent vector size (10 representations) on CK Database. Classifiers: single (SVM), Ensembles (BG: Bagging, RF: Random Forest), KnoraU (DT: pool of trees; RF: Random Forest). Fusion of all (sum, product and stacking)}
\begin{tabular}{l|l|ll|ll} \hline
\multicolumn{1}{c|}{} & \multicolumn{1}{c|}{} & \multicolumn{2}{c|}{Ensembles} &  \multicolumn{2}{c}{KnoraU}   \\                 
Repr.                & \multicolumn{1}{c|}{SVM}                     & \multicolumn{1}{c}{BG}                    & \multicolumn{1}{c|}{RF}                    & \multicolumn{1}{c}{DT} & \multicolumn{1}{c}{RF} \\ \hline
150      & 77.68                                    & 63.51                                   & 58.47                                   & 64.61                  & 59.42                  \\
200      & 84.13                                    & 63.10                                   & 63.54                                   & 63.31                  & 64.19                  \\
250      & 82.28                                    & 64.73                                   & 61.37                                   & 66.49                  & 64.12                  \\
300      & 83.95                                    & 63.02                                   & 58.88                                   & 64.60                  & 60.19                  \\
400      & 84.95                                    & 64.78                                   & 61.08                                   & 64.75                  & 60.70                  \\
500      & 87.06                                    & 71.07                                   & 63.37                                   & 71.52                  & 65.50                  \\
1000     & 86.46                                    & 68.82                                   & 66.34                                   & 67.55                  & 66.76                  \\
1500     & 86.54                                    & 67.76                                   & 64.15                                   & 68.94                  & 65.14                  \\
2000     & 86.96                                    & 71.96                                   & 65.28                                   & 72.59                  & 67.21                  \\
2500     & 85.97                                    & 68.33                                   & 63.10                                   & 67.06                  & 64.98                  \\ \hline
Sum      & 86.96                           & 75.46                                   & 66.97                                   & 75.29                  & 67.79                  \\
Product  & \textbf{86.99}                                    & 76.31                                   & 67.25                                   & 75.50                  & 67.79                  \\
Stacking & 86.29                                    & \textbf{78.79}                          & \textbf{74.30}                          & \textbf{76.49}         & \textbf{72.37}   \\ \hline     
\end{tabular}
 \label{table_hidden_CK}
\end{table}

\subsection{Experiment 5 - Combining all the representations}

Finally, we have combined all the 29 representations based on different seeds (10), CAE architectures (5), latent vector sizes (10), and auxiliary datasets (4). The results for JAFFE and CK datasets are shown in Tables \ref{tab_tecnicas} and \ref{tab_tecnicas_CK}. Concerning the JAFFE dataset, the best result (64.59\% of accuracy) was observed when using KnoraU combining the output of RFs. For the CK dataset, the best accuracy was 89.22\% observed when combining the output of the SVMs using the stacking strategy. It is well-known in the literature that the success of an ensemble or a dynamic selection method depends on a diverse pool of accurate classifiers. Thus, it may corroborate that the proposed strategies may generate diverse representations.

Tables \ref{state-of-art} and \ref{state-of-art-ck} show that most of the proposed strategies compare favorably against related works representing the state-of-the-art when applying similar unsupervised techniques for JAFFE and CK datasets. In the case of JAFFE, our best result (64.59\%) improved the state-of-art in 8.14 percentage points (from 56.45\% to 64.59\%). For CK, the area under the ROC curve (AUC) was computed to compare with the results reported in \cite{LONG}. We can observe in Table \ref{state-of-art-ck} that our results are better for most classes, losing only for \textit{sad}. Moreover, we considered the seven classes available in the CK dataset in our experiments.  

\begin{table}[htbp]
\centering
\caption{Results of Ensembles combining all strategies\\ (the different representations) on JAFFE Database}
\begin{tabular}{l|l|ll|ll} \hline
\multicolumn{1}{c|}{} & \multicolumn{1}{c|}{} & \multicolumn{2}{c|}{Ensembles} &  \multicolumn{2}{c}{KnoraU}   \\                 
Repr.                & \multicolumn{1}{c|}{SVM}                     & \multicolumn{1}{c}{BG}                    & \multicolumn{1}{c|}{RF}                    & \multicolumn{1}{c}{DT} & \multicolumn{1}{c}{RF} \\ \hline
Sum      & 60.63                & 62.72               & 61.78               & 62.68   & 64.15  \\
Product  & 61.10                & 63.22               & 61.24               & 63.61   & \textbf{64.59}  \\
Stacking & 60.33                & 60.29               & 62.24               & 61.22   & 63.10 \\ \hline
\end{tabular}
\label{tab_tecnicas}
\end{table}

\begin{table}[htbp]
\centering
\caption{Results of Ensembles combining all strategies\\ (the different representations) on CK Database}
\begin{tabular}{l|l|ll|ll} \hline
\multicolumn{1}{c|}{} & \multicolumn{1}{c|}{} & \multicolumn{2}{c|}{Ensembles} &  \multicolumn{2}{c}{KnoraU}   \\                 
Repr.                & \multicolumn{1}{c|}{SVM}                     & \multicolumn{1}{c}{BG}                    & \multicolumn{1}{c|}{RF}                    & \multicolumn{1}{c}{DT} & \multicolumn{1}{c}{RF} \\ \hline
Sum      & 86.05                & 74.29         & 67.14         & 73.77        & 67.07       \\
Product  & 86.27                & 75.98         & 66.93         & 75.04        & 67.92       \\
Stacking & \textbf{89.22}       & 77.28         & 71.14         & 75.18        & 77.72 \\ \hline
\end{tabular}
\label{tab_tecnicas_CK}
\end{table}

At this point, we can answer our two research questions. Answering RQ1, we can say that using a pool of unsupervised representations can contribute to the performance of the final classification process. Concerning RQ2, we have observed that the best approaches directly relate to the CAE architecture. Experiments have shown that varying the latent vector size (Jaffe experiments) and the number of CAE layers (CK experiments) were the most promising strategies. 

\begin{table}[htbp]
\centering
\caption{Comparison with related work also based on self-taught learning}
\begin{tabular}{l|l} \hline
\multicolumn{1}{c}{\textbf{Method}}   & \multicolumn{1}{c}{\textbf{Accuracy}} \\ \hline
Bhandari et al. \cite{BHANDARI}                      & 56.45                                 \\ \hline
Our strategy (single)                 &                                       \\
- SVM                                 & 60.35                                 \\
Our Strategies (ensembles)            &                                       \\
- Random Seeds                        & 56.20                                 \\
- Network Architecture                & 59.67                                 \\
- Different Unlabeled Datasets        & 61.69                                 \\
- Latent Vector                       & 62.26                                 \\ 
\textbf{- KNORA Fusion of Strategies} & \textbf{64.59}   \\ \hline                    
\end{tabular}
\label{state-of-art}
\end{table}

\begin{table*}[htbp]
\centering
\caption{Comparison with related work based on STL on CK Database (area under ROC curve).}
\begin{tabular}{lcccccccl} \hline
                                                                               & Anger          & Disgust        & Fear           & Joy            & Sad            & Surprise       & Contempt & Mean           \\ \hline
Long et al. \cite{LONG}                                                                           & 77.39          & 71.08          & 69.16          & 89.43          & \textbf{84.78} & 89.05          & ---      & 80.15          \\ \hline
\begin{tabular}[c]{@{}l@{}}Ours - SVM Stacking\\ (all strategies)\end{tabular} & \textbf{80.00} & \textbf{94.90} & \textbf{80.00} & \textbf{98.60} & 82.21          & \textbf{96.40} & 66.70    & \textbf{85.54} \\ \hline
\end{tabular}
\label{state-of-art-ck}
\end{table*}

\section{Conclusions and Future Work} \label{conclusion}

This paper introduced different strategies to generate unsupervised representations obtained with STL for FER. These strategies explore diversity by varying the autoencoders' initialization, architecture, and training data. In addition, support vector machines, bagging, random forest, and a dynamic ensemble selection method (KNORA) were evaluated as classifiers that learn from the unsupervised STL representations provided. 

Experimental results on two well-known FER datasets shows that the proposed strategies compare favorably when compared with related works representing the state-of-the-art when applying similar unsupervised representation learning strategies. By answering the proposed research questions we can say that using a pool of unsupervised representations can contribute to the performance of a supervised model. Moreover, a promising strategy to generate diversity is by using CAEs with different architectures, i.e., varying the latent vector size and the number of CAE layers. 

In future works, we intend to learn weights to combine the generated unsupervised representations. The idea is to optimize their fusion. In addition, we plan to apply the proposed strategies to other problems. 



\balance
\bibliographystyle{IEEEtran}

\balance

\end{document}